\documentclass{article}
\usepackage{spconf,amsmath,graphicx}
\usepackage{comment}
\usepackage{fullpage}
\usepackage{times}
\usepackage{fancyhdr,graphicx,amsmath,amssymb}
\usepackage[ruled,vlined]{algorithm2e}
\usepackage{scrextend}
\usepackage{todonotes}
\usepackage{xcolor}

\title{Incremental user embedding modeling for personalized text classification}
\name{Ruixue Lian$^{1\dagger}$, Che-Wei Huang$^2$, Yuqing Tang$^2$, Qilong Gu$^2$, Chengyuan Ma$^2$, Chenlei Guo$^2$ 
\thanks{$\dagger$Work was done as an intern at Amazon Alexa.}}
\address{$^1$University of Wisconsin-Madison, Dept. of Electrical and Computer Engineering \\ 
$^2$Amazon Alexa \\
ruixue.lian@wisc.edu, \{cheweh, yuqint, gqilon, mchengyu, guochenl\}@amazon.com}
\begin{document}
%
\maketitle
\begin{abstract}
Individual user profiles and interaction histories play a significant role in providing customized experiences in real-world applications such as chatbots, social media, retail, and education. Adaptive user representation learning by utilizing user personalized information has become increasingly challenging due to ever-growing history data. In this work, we propose an incremental user embedding modeling approach, in which embeddings of user's recent interaction histories are dynamically integrated into the accumulated history vectors via a transformer encoder. This modeling paradigm allows us to create generalized user representations in a consecutive manner and also alleviate the challenges of data management. We demonstrate the effectiveness of this approach by applying it to a personalized multi-class classification task based on the Reddit dataset, and achieve 9\% and 30\% relative improvement on prediction accuracy over a baseline system for two experiment settings through appropriate comment history encoding and task modeling.
\end{abstract}
\begin{keywords}
Personalized representation learning, incremental embedding, multi-class classification
\end{keywords}
\section{Introduction}
\label{sec:intro}
Given that archived user records such as user-specific demographic attributes and daily utterances can reveal users' personalization, interests and attitudes towards different topics, it allows us to infer user preferences and hence provide personalized user experiences based on the records collected by social media platforms or conversational agent systems. These context signals always reflect consistent personalities, which can provide meaningful information and thus can be leveraged to build coherent models to facilitate a variety of application needs better \cite{li2016persona,zhang2019consistent}. Thus, it becomes imperative to develop adaptive user representations by incorporating personalized features from different modalities for downstream tasks such as user preference prediction \cite{hasan2021learning, bui2019federated},  personalized dialogue generation \cite{wu2021personalized, zhang2018personalizing,pei2021cooperative}, and personalized recommendations \cite{grbovic2018real,kang2018self,zhou2018deep}.

Multiple previous work have been proposed to generate personalized representations, in which they assume the availability of users' entire history. For instance, Wu et al. built a personalized response generation system consists of a mechanism of splitting memories to capture personalization: one for user profile and the other for user-generated information such as comment histories \cite{wu2021personalized}. Grbovic et al. proposed a method capturing long-term and short-term user interests to build customized embeddings, which can be applied to real-time personalized recommendation \cite{grbovic2018real}. However, it is impracticable to access the entire historical data in some applications due to various constraints. To address this issue, Bui et al. proposed a way to utilize existing personalization techniques in the Federated Learning setting \cite{bui2019federated}. 

On the other hand, user embedding modeling in batch mode always train models from scratch by aggregating all available data. It is not only time-consuming but may also result in outdated models \cite{losing2018incremental}. Besides, it cannot capture the gradually-changing user interests and hence cannot reflect the recency of user data. To alleviate these issues, incremental learning has been proposed to update models in an iterative manner, which allows the model to learn new information as soon as it is available and thus leads to up-to-date models. It is not only capable of lifelong learning with restricted resources, but also captures the adaptive representation, and reduces the cost of data storage and maintenance \cite{bielak2021fildne,kabbach2019towards, wei2020incremental}.

Based on those findings, we propose an incremental user embedding modeling method to generate adaptive user representation by leveraging personalized information. We demonstrate the effectiveness of this approach on a multi-class text classification task on the Reddit dataset following \cite{bui2019federated} as this dataset could mimic real-world scenarios such as data collected by chatbots. One example sample is shown in Table \ref{table1}. The objective of this text classification task is to predict which class (i.e., subreddit) a given comment was posted by utilizing user profiles and comment histories, where user profiles refer to static user-specific key-value pairs. 

\begin{table}[h!]
\centering
\begin{tabular}{l}
\hline
\textbf{Incoming comment:} Watch funny anime.\\
\textbf{Subreddit:} Anime \\
\textbf{Author:} HaraJiang, \textbf{Time:} 2019-07-01 00:28:49 \\
\textbf{Comment histories:} \\
Victory Royale is really interesting. \\
Warframe is available for free, it is a good game. \\
I see a water dragon. \\
\hline
\end{tabular}
\caption{A constructed example from Reddit dataset} 
\label{table1}
\end{table}

\section{Proposed Methods}
\label{sec:methods}
This study focuses on building adaptive user representations in an incremental manner. In this section, we first introduce several batch learning methods to investigate model performance by utilizing personal information. Then we propose an incremental embedding learning method, which can continuously integrate newly available data into existing trained models and is effective in capturing the evolving change of user interests.

\begin{figure}[htp]
\centering
\centerline{\includegraphics[width=8cm]{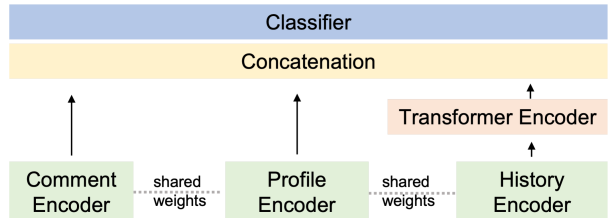}}
\caption{Architecture of the proposed model}
\end{figure}

\subsection{Batch Learning Methods}
We fine-tune a BERT model \cite{devlin-etal-2019-bert} for multi-class classification by adding a linear classifier layer on top of it. As shown in Fig. 1, the input is a concatenation of different types of feature representations, and the output is the predicted subreddit indicating where the given comment was posted. The model is trained and evaluated using three different input feature representations.

\begin{enumerate}
  \item $[q]$: incoming comment embedding $q$ only.
  \item $[q,U_p]$: concatenation of incoming comment embedding $q$ and static user profile representation $U_p$.
  \item $[q,U_p,U_h]$: concatenation of incoming comment embedding $q$, user profile representation $U_p$, and user history representation $U_h$.
\end{enumerate}

We also explore two ways to represent user history representation $U_h$. 

\begin{enumerate}
    \item \textbf{Mean-pooling}: use mean-pooling of all history embeddings as $U_h$.
    \item \textbf{Self-attention}: add an upper-layer transformer over all history embeddings and then use the mean-pooling of transformer output as $U_h$.
\end{enumerate}

\subsection{Incremental User Embedding Learning}


The core idea of incremental embedding learning method is  to dynamically integrate recent comment history embeddings into the accumulated history vectors via an upper-layer transformer encoder. As shown in algorithm 1, history embedding tensor $A$ and accumulated history embedding tensor $B$ are initialized first. Both $A$ and $B$ are in the shape of $n\times T\times d$, where $n$ is the number of users,  $T$ is the time-span of interaction history, and $d$ is the dimension of sequence embedding. $t_0$ is a timestamp chosen as history, and samples prior to $t_0$ are used for initialization. We then construct the dataset in the following form: $(i, t, y)$, where $t$ is a timestamp such that $t \in [t_0,T]$ and $y$ is the ground truth label of an incoming comment at $t$. This dataset was fed into the model in chronological order since $B$ is updated gradually.

For an incoming comment at $t$, the objective is to predict its associated subreddit by leveraging the accumulated history embedding in $B$. Meanwhile, $B$ is updated by integrating multiple latest comment history embeddings into it. Specifically, we get recent $n_0$ history embeddings and the accumulated history embedding from $A$ and $B$, respectively. Those vectors are input into the upper-layer transformer encoder and get attended with each other. U\_profile represents the personalized user profile incrementally constructed from comment histories. It is actually the mean-pooling of the transformer encoder output, and is used to update $B$. By using this momentum update, the impact of long-term history interactions is gradually decayed. Here, $\alpha$ is the momentum parameter, $\alpha \in [0,1]$.

With the proposed incremental user embedding learning method, accumulated user embeddings can be continuously updated with recent activities. It can adaptively capture the evolving change of user interests.


\begin{algorithm}
\SetAlgoLined
\textbf{Initialization:} \\
Let $i$ be user id, $i\in[0,n]$ \\
Let $j$ be history index, $j\in[0,T]$ \\
$A$[$i$][$j$] = encoder(history[$i$][$j$])  \\
$B$[$i$][$j$] = mean($A$[$i$][$:j]$) \\
\For{batch in dataset}{
    $q$ = encoder(history[$i$][$t$])\\
    history = $A$[$i$][$t-n_0: t-1$] \\
    past\_activity = $B$[$i$][$t-n_0-1$] \\
    U\_profile = mean(transformer([history, past\_activity])) \\
    $B$[$i$][$t-1$] = $\alpha$ * U\_profile + ($1 - \alpha$) * $B$[$i$][$t-1$] \\
    loss, logits = classifier([$q$, $B$[$i$][$t-1$]])
}

\caption{Incremental user embedding learning}
\end{algorithm}

\section{Experiments}
\label{sec:pagestyle}
\subsection{Datasets}
In this section, we describe how three datasets were built from the Reddit comment dataset\footnote{\label{1}https://files.pushshift.io/reddit/comments/} \cite{baumgartner2019reddit}. The dataset used to pre-train BERT model is referred to as ``pre-training set". The other two datasets used to investigate batch learning methods and incremental embedding learning are called ``subset1" and ``subset2", respectively.


The ``pre-training set" consists of comments from the top $256$ subreddits in $2019/07$ by following the procedure described in \cite{bui2019federated} . Each comment corresponds to one subreddit indicating where it was posted. There are around $5$M comments distributed in 256 classes in total, and this dataset is imbalanced.

The ``subset1" was randomly selected from the ``pre-training set" by balancing the subreddit distribution. User profiles (e.g. Username and created time associated with each comment) are also kept. In addition, we pick up 5 comment histories that are most recent to each incoming comment and we list them in the reverse chronological order. There are around 126K incoming comments in total. We split them into training, validation, and test sets by 50\%, 25\%, and 25\%.

The ``subset2" was built by picking up 40 histories for each incoming comment. Besides, history's associated subreddits are also kept, while two user-related attributes are thrown away. The other settings remain the same as subset1. After customizing the sequential dataset as mentioned in section 2.2, the subset2 distribution looks similar to the original imbalanced one.

\subsection{Experimental Setup}
All three datasets were preprocessed by standard text normalization steps, such as removing urls, trimming extra spaces, and filtering non-ASCII words and characters. A BERT model with shared weights is used as encoder, and its output of the first token ([CLS] token) is used as sequence embedding. 

To conduct personalized prediction on subset1, the BERT model is fine-tuned together with other hyperparameters. User profiles are represented as key-value pairs, which are concatenated into the input sequence before encoding. 

In incremental embedding modeling on subset2, we first use masked language model (MLM) to pre-train BERT with and without history's associated subreddit, respectively. To include subreddits of history data as input, we first expand tokenizer vocabulary. Then we prepend subreddits in textual format in front of history sequences. After pretraining, we extract history embeddings from frozen BERT with different strategies depending on whether we prepend their associated subreddits or not. We weight the cross-entropy loss by the inverse frequency of each category to deal with the imbalanced dataset.

We use the bert-base-uncased model provided in huggingface’s transformer package \cite{wolf2019huggingface}. For the upper-layer transformer, we use 4 layers of transformer encoder, and each layer with 8 attention heads. AdamW is used as our optimizer with an initial learning rate of 2e-5 and we use a linear decay learning rate scheduler across all model training. BERT max sequence length is 128 in all settings. The dropout probability is always 0.1. For pre-training, we use batch size of 16 and 5 epochs. For all other experiments, we use batch size of \{64, 512, 1024\}. The maximum number of iteration is set as 15 with an early stop if no improvement over accuracy on validation set. $\alpha$ is 0.1 for incremental update. It takes no more than 40 mins to perform incremental modeling with 8 NVIDIA V100 GPUs.

\subsection{Experimental Results and Analysis}
Table \ref{table2} summarizes the classification accuracy of batch learning methods described in section 2.1. We observed that: (1) the network incorporates user profile or comment histories can boost the model performance. The classification accuracy is improved by 7\% relatively by incorporating user profiles, and 41\% relatively with the mean-pooling of history embeddings as additional feature; (2) the attention between posting histories contributes 62\% relative improvement to accuracy. It shows that self-attention between histories would enhance personalized features and reinforce users' preference and interests. Although the model with self-attention over all history embedding performs better compared to using history embedding mean-pooling, it is still a trade-off between these two methods in terms of data storage capacity and computational resources since model with self-attention has more complicated structure, leading to more hyperparameters.

\begin{table}[h!]
\centering
\begin{tabular}{c c c c}
\hline
Model & Feature & Acc(\%) & Rel.Imp.(\%) \\
\hline
BERT & [$q$] & 17.95 & 0\\
BERT & [$q, U_p$] & 19.23 & 7  \\
Mean-pooling & [$q, U_p, U_h$] & 25.37 & 41 \\
Self-attention & [$q, U_p, U_h$] & 29.02 & 62 \\
\hline
\end{tabular}
\caption{Batch learning method performance.}
\label{table2}
\end{table}

The results of the mean-pooling method are regarded as the lower bound. Similarly, the results with upper-layer transformer encoder are regarded as an upper bound since attention behaviors over entire histories enhance personalized semantic features. The goal of incremental embedding modeling is to  asymptotically approximate the upper bound method and get their performance as close as possible. Moreover, the incremental modeling would capture the recency of user data and reflect the evolving change of user interests over time while these cannot be achieved by self-attention method. We do not include static $U_p$ in incremental update since it does not play a significant role as results shown in Table \ref{table2}.

Table \ref{table3} shows the incremental modeling results. It cannot be directly compared with Table \ref{table2} since it is performed on sequential dataset instead. Histories are encoded without and with considering their associated subreddits in the 2nd and 3rd column respectively. From 2nd column, our incremental method lies in between the lower and upper bounds as we expected. It improves the lower bound mean-pooling method by 9\% relatively, and is 14\% relatively below the upper bound self-attention approach in terms of accuracy.

In the 3rd column of Table \ref{table3}, history's associated subreddits are considered as part of history attributes and are prepended in the texual format in front of histories before encoding. In this way, both the prior information and the semantics of history subreddits are incorporated into the sequence embedding. From the results, the accuracy of proposed incremental modeling is 30\% relatively higher than the mean-pooling method, and is only 2\% behind the upper bound. It significantly closes the gap between the incremental modeling and the self-attention method compared to 2nd column. This is because the attention behaviors introduced by the upper-layer transformer reinforce the personalised features, especially the prior information, which provides the most direct information in classification. 

 By comparing the incremental modeling performances with two ways of history encoding, around 59\% relative improvement in accuracy can be achieved. To further investigate the performance gain introduced by prior, we predict the incoming comment by the majority count of its history's subreddits as a direct comparison. The incremental modeling can improve the majority count by 38\% in accuracy. These analyses indicate that the incremental modeling effectively utilizes semantics from both utterances and history prior.

\begin{table}[h!]
\centering
\begin{tabular}{c c c c}
\hline
Model  & Acc (\%) & Acc (\%)\\
       & w/o subreddit & w subreddit \\
\hline
Majority count & - & 36.06 \\ 
Mean-pooling & 28.79 & 38.27\\
Self-attention & 36.72 & 50.80\\
\textbf{Incremental} & \textbf{31.44} & \textbf{49.94}\\
\hline
\end{tabular}
\caption{Incremental embedding modeling performance. Histories are encoded without and with their associated subreddits in 2nd and 3rd column, respectively.} 
\label{table3}
\end{table}

\section{Conclusions and Future Work}
\label{sec:majhead}
In this work, we proposed a method to generate personalized adaptive user representations by utilizing user profiles and user interaction histories in a consecutive manner. Specifically, we proposed an incremental user embedding modeling paradigm, which can dynamically integrate most recent user activities into the accumulated history embedding vectors. We show that a better performance can be achieved on downstream predictions with proper history encoding. Besides,  we show that this approach not only captures the recency of user data and reflects the evolving change of user interests, but also keeps the model up-to-date and improves data storage efficiency.

Since the old accumulated history vectors do not update in a timely manner with the updates of the model, we have to introduce other methods such as regularization to improve the performance in the future. Additionally, we would like to extend the proposed incremental modeling to other applications such as personalized intent classification and personalized recommendation.




\bibliographystyle{IEEEbib}
\bibliography{main}

\end{document}